\documentclass[a4paper, 10pt, conference]{ieeeconf}      
                                                          
\usepackage{FG2020}

\FGfinalcopy

\IEEEoverridecommandlockouts

\usepackage{graphicx}
\usepackage {subfig}

\def\FGPaperID{****} 

\title{\LARGE \bf
Action Units Recognition by Pairwise Deep Architecture
}

\author{\parbox{16cm}{\centering
    {\large Junya Saito, Ryosuke Kawamura, Akiyoshi Uchida, Sachihiro Youoku,\\ Yuushi Toyoda, Takahisa Yamamoto, Xiaoyu Mi and Kentaro Murase}\\
    {\normalsize Artificial Intelligence Laboratory, Fujitsu Laboratories Ltd., Kanagawa, Japan}}
}

\begin{document}

\ifFGfinal
\thispagestyle{empty}
\pagestyle{empty}
\else
\author{Anonymous FG2020 submission\\ Paper ID \FGPaperID \\}
\pagestyle{plain}
\fi
\maketitle

\begin{abstract}

In this paper, we propose a new automatic Action Units (AUs) recognition method used in a competition, Affective Behavior Analysis in-the-wild (ABAW).
Our method tackles a problem of AUs label inconsistency among subjects by using pairwise deep architecture.
While the baseline score is 0.31, our method achieved 0.67 in validation dataset of the competition.

\end{abstract}

\section{INTRODUCTION}

Automatic Action Units (AUs) recognition is useful and important in facial expression analysis~\cite{zhi2020comprehensive, martinez2017automatic}.
AUs represent the muscular activity that produces momentary changes in facial appearance and for example AU4 indicates brow lowerer and AU6 indicates cheek raiser and lid compressor~\cite{ekman2002facial}.
AUs are scored by occurrence or intensity.
AUs occurrence is described by binary scale and AUs intensity is described by neutral or five-point ordinal scale, A-B-C-D-E, where E refers to maximum evidence.
AUs occurrence or intensity (AUs label) is determined by human experts, called as coders, based on facial appearance change of target subjects.

At this time, a competition including automatic AUs recognition task, Affective Behavior Analysis in-the-wild (ABAW), was held in FG2020~\cite{kollias2020analysing,kollias2019expression,kollias2018aff,kollias2018multi,kollias2019deep,zafeiriou2017aff,kollias2017recognition}.
In the competition, training and validation datasets that include multiple videos and AUs occurrence annotation for each frame image of the videos are provided.
Participants are required to submit AUs occurrence recognition results for each frame image of test dataset videos and are compared based on an evaluation metric composed of F1 and accuracy.
In this paper, we explain our new automatic AUs recognition method used in the competition.

AUs scoring is defined for facial appearance change.
Besides, it is said that the appearance change varies by subjects depending upon their bone structure, variations in the facial musculature, permanent wrinkles, etc. in FACS manual~\cite{ekman2002facial}.
This means that AUs label criteria for facial appearance change might be inconsistent in different subject's videos.
This inconsistency might cause a problem to degrade a performance of simple method that predicts AUs label from only single image~\cite{niinuma2019unmasking}.

To tackle the problem, we propose a method based on the following assumption about coders' scoring process. Coders first observe the whole video of a target subject and understand variation of facial appearance change in the video. Then coders infer a mapping from degree of facial appearance change to AUs intensity based on the variation range.

To follow the process of understanding the variation, we introduce pseudo-intensity which represents subject-independent degree of facial appearance change, for example degree of  inner blow movement or wrinkle depth in the case of AU4,  and we train a model to output pseudo-intensity by pairwise deep architecture like siamese network~\cite{doughty2018s}.
And then we train a mapping model to convert pseudo-intensity to AUs label based on the variation of pseudo-intensities in the video.

\section{RELATED WORKS}
In this section we introduce related previous methods and explain relation to our method.

\subsection{Methods using temporal features}

Tadas et al. proposed a method to normalize feature vector by median of temporally varying feature vector in target video~\cite{baltruvsaitis2015cross} and a method to normalize recognition result value by n-th percentile of temporally varying recognition result value~\cite{baltruvsaitis2016openface}.
These methods can capture neutral face by using median or n-th percentile but cannot capture variation range of facial appearance change.

Jun et al. and Wen-Sheng et al. proposed a method to be able to capture temporal features by using RNN or LSTM~\cite{he2017multi,chu2017learning}.
If we set sequence length long enough and employ temporally bidirectional network, specifically bidirectional-LSTM, and train with a sufficiently large amount of AUs dataset, then the RNN approach conceptually may be able to solve a problem of AUs label criterion inconsistency.
Each method however does not satisfy both condition of sequence length and employment of bidirectional network, and it is difficult to prepare large amount of AUs dataset because coders' annotation takes a very long time~\cite{martinez2017automatic}.
Thus, it is practically difficult to solve the problem by the RNN approach.

\begin{figure*}
  \centering
  \includegraphics[width=11cm]{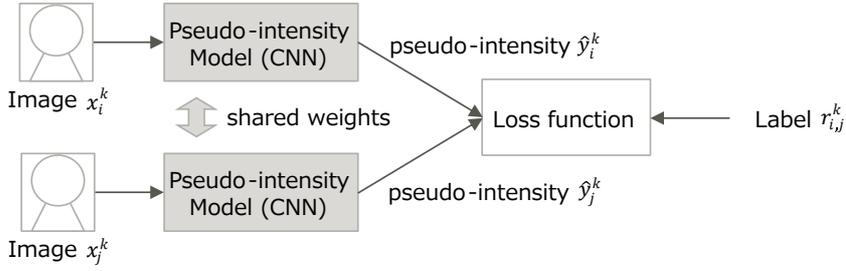}
  \caption{Pairwise deep architecture for training pseudo-intensity model}
  \label{pairwise_network}
\end{figure*}

\subsection{Methods using pairwise architecture}

Tadas et al. and Paul Pu et al. proposed a method to calculate pseudo-intensity based on global ranking in a target video
and to convert pseudo-intensity to label by using normalization function or RNN~\cite{baltruvsaitis2017local,liang2018multimodal}.
The global ranking is calculated by merging local rankings in the target video. The local ranking indicates an intensity ranking of image pair.
The normalization function is similar to our mapping model but does not capture variation range of facial appearance change and there are some problems in RNN approach as mentioned above.
Thus, it is difficult to solve a problem of AUs label criterion inconsistency by these methods.
Additionally, their pseudo-intensity is based on only comparative relationship in target video,
thus the method does not perform well for video of which intensity is high stationarily.
Besides, our pseudo-intensity is subject-independent, thus our method is expected to perform well for the case.

\section{METHODOLOGY}

\begin{figure*}
  \centering
  \includegraphics[width=15cm]{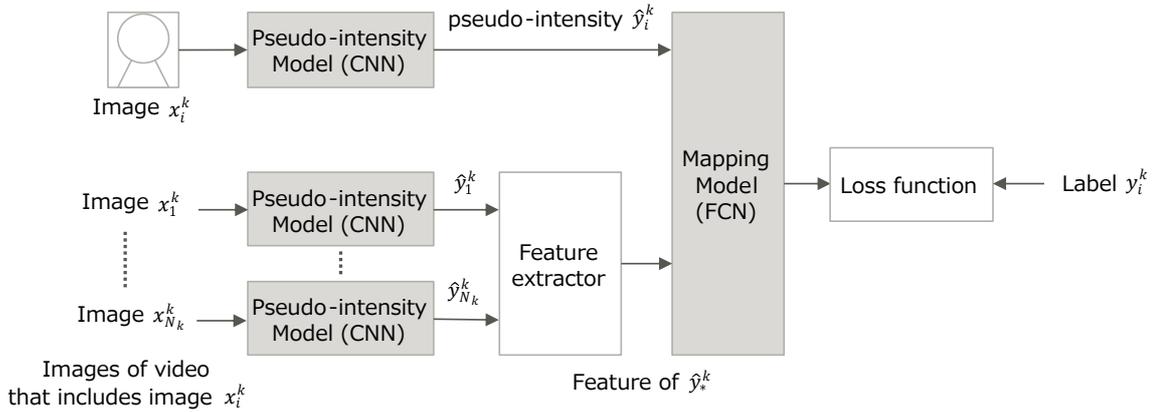}
  \caption{Architecture for training mapping model}
  \label{normalization_function}
\end{figure*}

In this section we explain our new method for automatic AUs recognition to tackle a problem of AUs label inconsistency.

The method consists of two steps in training phase.
In training phase, first, we train a model to output pseudo-intensity, that represents degree of facial appearance change.
Training dataset for the model consists of image pairs in same video with labels created from videos of various subjects.
The label is AUs intensity ranking of the image pair.
The model is trained to make pseudo-intensity ranking and intensity ranking equal by using the training dataset and pairwise deep architecture
on the basis that AUs label criteria for facial appearance change is consistent in same video.

Second, we train a mapping model to convert pseudo-intensity to AUs label based on variation range of pseudo-intensities in a video.
In this paper, variation range feature is composed of percentile and frequency feature of pseudo-intensities in the video.
In predicting phase, the method receives target image and frame images of video including the target image and calculates pseudo-intensities and converts to AU label by using the mapping model.

In the rest of this section, we explain the detail of training phase.

\subsection*{Step1. Training Pseudo-intensity Model}

Given a set of $n$ input images with their corresponding labels $\{ (x_i^k \in \mathcal{R}^d); (y_i^k \in [0,5] ) \}^n$, where $k$ is subject id and $i$ is video frame id of the subject,
we construct training dataset $\{(x_i^k,x_j^k); (r_{ij}^k \in \{0,1\})\}^m$ of size $m$ for pseudo-intensity model. we define $r_{ij}^k$ as:
\begin{equation}
r_{ij}^k = \left\{
\begin{array}{ll}
1 & \mbox{if } y_i^k < y_j^k \\
0 & \mbox{if } y_i ^k > y_j^k
\end{array}
\right.
\end{equation}
The training dataset is made by sampling from a set of input images and labels.

We next construct pairwise deep architecture for training pseudo-intensity model as Fig.\ref{pairwise_network}.
Let the model be Convolutional Neural Network (CNN).
The model gets image $x_i^k$ and outputs pseudo-intensity $\hat{y}_i^k \in \mathcal{R}$ and another model of which weight is shared gets image $x_j^k$ and outputs pseudo-intensity $\hat{y}_j^k \in \mathcal{R}$. A loss function consists of $\hat{y}_i^k$, $\hat{y}_j^k$ and  $r_{ij}^k$ as:
\begin{equation}
\sum_{\hat{y}_i^k, \hat{y}_j^k, r_{ij}^k}{\max{(0,t - (1-2r_{ij}^k) ( \hat{y}_i^k - \hat{y}_j^k ))}},
\end{equation}
where $t=1$.
The loss function is similar in \cite{doughty2018s}.
The pseudo-intensity model is trained by the training dataset and the pairwise deep architecture.

\subsection*{Step2. Training Mapping Model}

We generate pseudo-intensities by using trained pseudo-intensity model and train a mapping model by the pseudo-intensities and AUs label.
We present an architecture for training mapping model at Fig.\ref{normalization_function}.
Let training dataset for mapping model be $ \{ (\hat{y}_i^k, G(\{ \hat{y}_1^k, ..., \hat{y}_{N_k}^k \}) ); y_i^k \}^l$ of size $l$, which $N_k$ is number of video frames of subject $k$ and $G$ is feature extractor of pseudo-intensities of video of subject $k$.
The feature extractor $G$ generates variation range feature of the pseudo-intensities. Specifically, the feature is percentile feature (0-th percentile, 10-th percentile, ...) and frequency feature (frequency in range $[ f_0, f_1 ]$, frequency in range $[ f_1, f_2 ]$, ...).
The training dataset is made by sampling from a set of input images and labels.

We compose the mapping model by Fully Connected Network (FCN). Let a loss function be cross entropy loss.

\section{EXPERIMENT}

In this section we explain a experiment result using the competition dataset.

\subsection{Datasets}
We used a dataset provided in the competition, called as Aff-Wild2, and two additional datasets.
The additional datasets are BP4D~\cite{zhang2014bp4d,zhang2013high} with AUs intensity and DISFA~\cite{mavadati2013disfa,mavadati2012automatic} with AU intensity.
From BP4D, images of nine different face orientations were created in FERA2017~\cite{valstar2017fera} and we used it.
Moreover, we created images of additional different face orientations, that is 60 and 80 degrees yaw and we created mirrored images of these and we used it.
About DISFA, we used original images.
As training dataset for pseudo-intensity model, we used BP4D with AU intensity and DISFA with AU intensity and Aff-Wild2 with AU occurrence.
As training dataset for mapping model, we used Aff-Wild2 with AU occurrence.
As validation dataset, we used Aff-Wild2 with AU occurrence.

\subsection{Settings}
The CNN of pseudo-intensity model is configured with VGG16 network pre-trained on ImageNet~\cite{simonyan2015very} and the FCN of mapping model is configured as classifier layer of VGG16.
As pre-processing, we applied procrustes analysis for images according to~\cite{niinuma2019unmasking}.
We selected best result based on validation dataset score in 5 training trials of same conditions because a randomness at initialization or training process may change performance.

\subsection{Evaluation Metric}

In the competition, an evaluation metric is defined. The metric is:
\begin{equation}
0.5 \times \mbox{F1\_score} + 0.5 \times \mbox{Accuracy},
\end{equation}
where F1 score is the unweighted mean and Accuracy is the total accuracy.

\subsection{Result}

Table~\ref{result_score} presents results of baseline and our method in validation dataset.
The baseline is in ~\cite{kollias2020analysing}.
Test dataset is not released, thus we evaluated by validation dataset.
The result indicates that our method outperforms the baseline.
It however does not represent that our assumption is correct or our method performs as expected.
The analysis about this is our future work.

\begin{table}
\caption{Results on validation dataset}
\label{result_score}
\begin{center}
\begin{tabular}{|c|c|c|c|}
\hline
 & Average F1& Total Accuracy & Competition Metric \\
\hline
Baseline~\cite{kollias2020analysing} & 0.22 & 0.4 & 0.31 \\
\hline
Ours & 0.39 & 0.95 & 0.67 \\
\hline
\end{tabular}
\end{center}
\end{table}

\section{CONCLUSION}

We proposed a new automatic AUs recognition method used in a competition, ABAW.
Our method uses pairwise deep architecture to tackle a problem of AUs label inconsistency among subjects.
Moreover, we compared our method and a baseline in the competition evaluation metric, and the result presented our method outperforms the baseline.
As future work, we will analysis that our assumption is correct and our method performs as expected.

\bibliographystyle{ieeetr}
\bibliography{sample}

\end{document}